\documentclass [sigconf]{acmart}
\AtBeginDocument{%
  }

\copyrightyear{2026}
\acmYear{2026}
\setcopyright{cc}
\setcctype{by}
\acmConference[WSDM '26]{Proceedings of the Nineteenth ACM International Conference on Web Search and Data Mining}{February 22--26, 2026}{Boise, ID, USA}
\acmBooktitle{Proceedings of the Nineteenth ACM International Conference on Web Search and Data Mining (WSDM '26), February 22--26, 2026, Boise, ID, USA}
\acmPrice{}
\acmDOI{10.1145/3773966.3778011}
\acmISBN{979-8-4007-2292-9/2026/02}

\usepackage{multirow}

\begin{document}

\title{Breaking Determinism: Stochastic Modeling for Reliable Off-Policy Evaluation in Ad Auctions}

\author{Hongseon Yeom}
\authornote{Both authors contributed equally to this research.}
\orcid{0009-0001-8866-2580}
\affiliation{%
  \institution{NAVER Corporation}
  \city{Seongnam-si}
  \country{Republic of Korea}
}
\email{yeom.j.hs@navercorp.com}

\author{Jaeyoul Shin}
\authornotemark[1]
\orcid{0009-0004-0274-7148}
\affiliation{%
  \institution{NAVER Corporation}
  \city{Seongnam-si}
  \country{Republic of Korea}
}
\email{jaeyoul.shin@navercorp.com}

\author{Soojin Min}
\orcid{0009-0009-0671-3419}
\affiliation{%
  \institution{NAVER Corporation}
  \city{Seongnam-si}
  \country{Republic of Korea}
}
\email{soojin.min@navercorp.com}

\author{Jeongmin Yoon}
\orcid{0009-0005-2889-2500}
\affiliation{%
  \institution{NAVER Corporation}
  \city{Seongnam-si}
  \country{Republic of Korea}
}
\email{jeongmin.yoon@navercorp.com}

\author{Seunghak Yu}
\orcid{0000-0003-0480-3110}
\affiliation{%
  \institution{NAVER Search US}
  \city{Seattle}
  \state{Washington}
  \country{United States}
}
\email{seunghak.yu@navercorp.com}

\author{Dongyeop Kang}
\orcid{0000-0002-9021-1789}
\authornote{Corresponding author.}
\affiliation{%
  \institution{NAVER Search US, University of Minnesota}
  \country{United States}
}

\email{dongyeop@umn.edu}

\renewcommand{\shortauthors}{Hongseon Yeom et al.}

\newcommand{\dk}[1]{\textcolor{teal}{\bf\small [#1 --DK]}}
\newcommand{\jy}[1]{\textcolor{orange!80!black}{\textbf{\small [#1 --JY]}}}
\newcommand{\hs}[1]{\textcolor{green!80!black}{\textbf{\small [#1 --HS]}}}
\newcommand{\sk}[1]{\textcolor{cyan!80!black}{\textbf{\small [#1 --SH]}}}
\begin{abstract}
Online A/B testing, the gold standard for evaluating new advertising policies, consumes substantial engineering resources and risks significant revenue loss from deploying underperforming variations. This motivates the use of Off-Policy Evaluation (OPE) for rapid, offline assessment. However, applying OPE to ad auctions is fundamentally more challenging than in domains like recommender systems, where stochastic policies are common. In online ad auctions, it is common for the highest-bidding ad to win the impression, resulting in a deterministic, winner-takes-all setting. This results in zero probability of exposure for non-winning ads, rendering standard OPE estimators inapplicable.
We introduce the first principled framework for OPE in deterministic auctions by repurposing the bid landscape model to approximate the propensity score.
This model allows us to derive robust approximate propensity scores, enabling the use of stable estimators like Self-Normalized Inverse Propensity Scoring (SNIPS) for counterfactual evaluation.
We validate our approach on the AuctionNet simulation benchmark and against 2-weeks online A/B test from a large-scale industrial platform. Our method shows remarkable alignment with online results, achieving a 92\% Mean Directional Accuracy (MDA) in CTR prediction, significantly outperforming the parametric baseline. MDA is the most critical metric for guiding deployment decisions, as it reflects the ability to correctly predict whether a new model will improve or harm performance. This work contributes the first practical and validated framework for reliable OPE in deterministic auction environments, offering an efficient alternative to costly and risky online experiments.
\end{abstract}




\begin{teaserfigure}
\centering
  \includegraphics[trim=0cm 0cm 0cm 0cm,clip,width=0.8\textwidth]{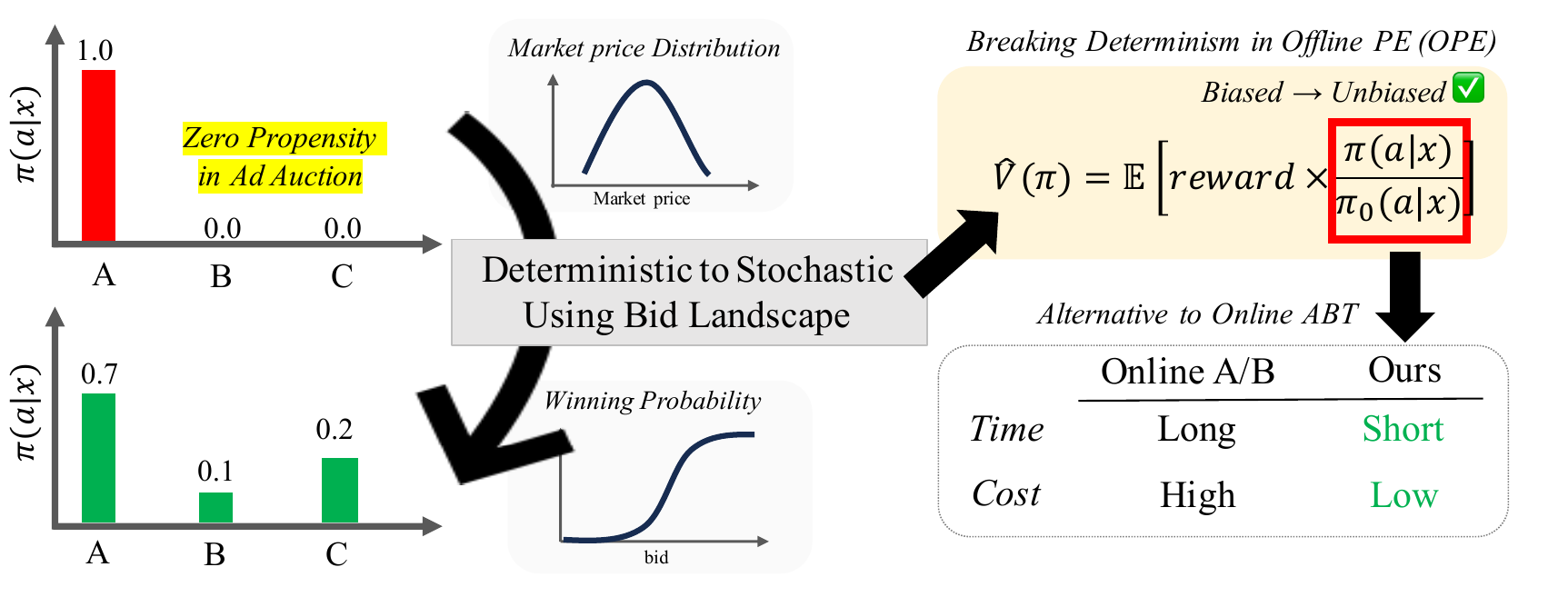}
  \vspace{-4mm}
  \caption{Breaking determinism in offline policy estimation in Ad auction via bid landscape modeling.}
  \Description{Conceptual framework of our proposed method. In deterministic ad auctions, only the winning ad has a non-zero propensity, making standard Off-Policy Evaluation (OPE) inapplicable (left). We leverage bid landscape modeling to transform the deterministic exposure into a stochastic distribution of winning probabilities (center). This enables the use of OPE, providing a fast and low-cost alternative to online A/B testing for evaluating new policies (right).}
  \label{fig:teaser}
\end{teaserfigure}


\maketitle

\section{Introduction}
Online A/B testing is the gold standard for evaluating new policies in large scale web systems, from user interface changes to new algorithmic models. However, this gold standard comes at a steep price. Deploying a new policy, even for a limited test, requires significant engineering resources and time. More critically, it carries the inherent risk of exposing users to an underperforming model, which can directly harm key business metrics such as user engagement and revenue. The ability to reliably predict a policy's performance \textit{offline}, before deployment, is therefore not just a matter of efficiency, but a crucial component of responsible and cost-effective innovation.

Off-Policy Evaluation (OPE) offers a compelling alternative by estimating the value of a new policy using historical data collected under a different policy. While OPE has achieved notable success in domains such as recommender systems~\cite{gilotte2018offline, narita2021debiased, saito2024long}, its application to online advertising auctions presents unique challenges. Recommender systems often employ stochastic policies, where multiple items have a non-zero probability of being displayed, enabling computation of the propensity scores required by standard estimators such as Inverse Propensity Scoring (IPS). In contrast, real-time bidding (RTB) ad auctions employ a deterministic, winner-takes-all mechanism: for a given ad request, only the highest-bidding ad is shown. This creates what we term the \emph{curse of the ad auction}: the observed propensity of any non-winning ad being shown is exactly zero, rendering standard OPE estimators mathematically inapplicable.

\textbf{Limitations of existing methods}.
A common workaround in the deterministic OPE literature is to \emph{relax} the policy using kernel-based smoothing~\cite{kallus2020doubly, lee2022local, lee2024kernel}, artificially introducing stochasticity by assuming a continuous probability distribution (e.g., a Gaussian kernel) around the chosen action.  However, these generic methods are ill-suited for the unique characteristics of ad auctions. First, the action space selecting one winner from N discrete ad candidates lacks the natural distance metric that continuous kernels rely on. Second, and more importantly, such methods ignore the complex, often multimodal nature of the underlying market price distributions and face severe computational scalability challenges when applied to the winner determination problem in auctions.

In this paper, we propose a novel framework that sidesteps these limitations by leveraging the inherent structure of the auction mechanism. 
Rather than modeling the discrete, deterministic action (which ad to show), we focus on the continuous, stochastic variable governing the outcome: the \emph{market price} (i.e., the second-highest bid). 
To this end, we adapt the Discrete Price Model (DPM), a technique from the bid landscape forecasting literature. 
While DPM was originally developed for bid optimization, we demonstrate its novel and powerful application for policy evaluation. By modeling the distribution of market prices, our framework can compute the winning probability for any given ad's bid. This probability serves as a principled and effective Approximate Propensity Score (APS), enabling the use of standard OPE estimators, otherwise intractable setting. Figure 1 provides an overview of the process.

\textbf{Contributions}.
First, to the best of our knowledge, this is the first attempt to introduce a practical and validated framework for applying OPE in deterministic, winner-takes-all ad auction environments. Second, we demonstrate the novel use of the Discrete Price Model (DPM) to estimate effective propensity scores, providing a domain-aware, computationally tractable alternative to generic smoothing methods. Third, we rigorously validate our framework on both the AuctionNet simulation benchmark and a 2-week online A/B test in a large scale industrial platform, achieving 92\% mean directional accuracy (MDA) in CTR prediction. Our results indicate that this offline evaluation framework can closely replicate online experiment outcomes, substantially reducing the cost and risk of model evaluation in advertising setups.

\section{Related Works}

\subsection{Off-Policy Evaluation}
Off-Policy Evaluation (OPE) estimates the value of a target policy $\pi$ using logs generated by a different behavior policy $\pi_0$~\cite{uehara2022review, voloshin2019empirical}. Importance sampling (IS) methods such as Inverse Propensity Scoring (IPS)~\cite{horvitz1952generalization} and its self-normalized variant SNIPS~\cite{swaminathan2015self} correct for distributional shift via importance weights $w(x,a)=\frac{\pi(a|x)}{\pi_0(a|x)}$.  
These estimators rely on the \emph{common support} assumption~\cite{narita2023counterfactual}, requiring that any action chosen by $\pi$ has non-zero probability under $\pi_0$. Deterministic auctions inherently violate this condition, which is the central challenge our work addresses.



\subsection{OPE in Recommender Systems and Advertising Auctions}
OPE has been extensively applied in recommender systems~\cite{saito2021evaluating, swaminathan2017off}, where stochastic logging, such as $\epsilon$-greedy exploration or randomization of item order, ensures that most items retain a non-zero probability of exposure. This inherent stochasticity satisfies the common support assumption and enables direct use of IPS or Doubly Robust (DR) estimators, though the large action spaces can still induce high variance~\cite{saito2021evaluating, swaminathan2017off}.

In contrast, real-time bidding (RTB) ad auctions are inherently deterministic and \emph{winner-takes-all}: the highest bidder wins with probability one, and all others have zero probability of being shown~\cite{bottou2013counterfactual, mohri2016learning}. This zero-propensity phenomenon means that standard IPS-based estimators cannot evaluate counterfactual outcomes for losing ads.

Prior work on OPE in advertising has focused on orthogonal challenges. For example, delayed feedback in conversion events has been addressed through specialized modeling~\cite{chapelle2011empirical}, and Sagtani et al.~\cite{sagtani2024ad} studied OPE for ad-load balancing, leveraging randomized logging policies to ensure full support. While effective in their respective contexts, these methods rely on policy randomization, which is often infeasible in production RTB systems due to business or performance constraints. As a result, a gap remains for approaches that operate directly on deterministic logs without altering the serving policy.




\subsection{Bid Landscape Forecasting}
Bid Landscape Forecasting (BLF) predicts the market price distribution $P(z|x)$ to optimize bidding strategies~\cite{ren2019deep, ghosh2019scalable}. The task is complicated by censored data when losing bids reveal only that $z$ exceeded the submitted price. Models such as DLF~\cite{ren2019deep} and ADM~\cite{li2022arbitrary} address this for bid optimization.  
Our work repurposes BLF for OPE: instead of selecting optimal bids, we use the market price distribution to compute the probability of winning for a given bid, yielding an \emph{Approximate Propensity Score} (APS). This bridges BLF and OPE, providing a principled way to overcome the zero propensity barrier in deterministic auctions.

\section{Problem Formulation}

Off-Policy Evaluation (OPE) offers a promising alternative to costly online A/B tests by leveraging historical log data. However, applying standard OPE methods to deterministic environments like real-world ad auctions presents a fundamental challenge. This section formally defines the problem, illustrating why the deterministic, winner-takes-all nature of ad auctions systematically biases OPE results, motivating our need for a new framework.

\subsection{Preliminaries}

To formally ground our discussion, we first define the key components in the context of both OPE and online advertising auctions.

\paragraph{Off-Policy Evaluation.} The primary goal of OPE is to estimate the expected outcome, or value $V(\pi)$, of a new evaluation policy $\pi$ using a historical log dataset $\mathcal{D} = \{(x_i, a_i, r_i)\}_{i=1}^n$. This dataset was collected by a different logging policy $\pi_0$. The value of a policy is the expected reward (e.g., Click-Through Rate, CTR) it would achieve if deployed:
\begin{equation}
    V(\pi) = \mathbb{E}_{a \sim \pi( \cdot | x), x \sim p(x)} [r(x, a)]
\end{equation}

The most fundamental OPE estimator is the Inverse Propensity Scoring (IPS) estimator \cite{horvitz1952generalization, bottou2013counterfactual, gilotte2018offline}, which re-weights the observed rewards by the ratio of propensity scores between the two policies:
\begin{equation}
    \hat{V}_{IPS}(\pi) = \frac{1}{n} \sum_{i=1}^n \frac{\pi(a_i|x_i)}{\pi_0(a_i|x_i)} r_i(x_i, a_i)
\end{equation}
Here, the ratio $w_i = \frac{\pi(a_i|x_i)}{\pi_0(a_i|x_i)}$ is the importance weight. For this estimator to be unbiased, a critical assumption known as common support is required: if an action $a$ could be taken by the evaluation policy ($\pi(a|x) > 0$), it must also have a non-zero probability of being taken by the logging policy ($\pi_0(a|x) > 0$).

\paragraph{Online Ad Auctions.}

\begin{itemize}
    \item \textbf{Context ($x$)}: A feature vector representing a user, a webpage, and other situational information for an ad impression; the vector can contain both discrete and continuous features.
    \item \textbf{Action ($a$)}: The specific ad creative chosen to be displayed for the given context $x$. The set of available actions is the pool of candidate ads, $A(x)$.
    \item \textbf{Reward ($r$)}: An indicator of a click on the displayed ad ($r=1$ for a click, $r=0$ otherwise).
    \item \textbf{Policy ($\pi$)}: A function that, given a context $x$, selects an ad $a$ from the set of candidates $A(x)$. In modern ad systems, this selection is based on a score, often a product of predicted CTR (pCTR) and a bid price.
\end{itemize}

\subsection{Limitation of OPE in Deterministic World}

The core challenge arises because the logging policy in a real-world ad auction is not stochastic; it is \textbf{deterministic}. The system deterministically selects the single ad with the highest score.

Let $\text{score}(a, x)$ be the score assigned to an ad $a$ for context $x$. The logging policy $\pi_0$ is an `argmax' function:
\begin{equation}
    \pi_0(x) = a^* = \underset{a \in A(x)}{\text{argmax}} (\text{score}_0(a, x))
\end{equation}
This means the true probability of an ad being shown is either 1 or 0:
\begin{equation}
    \pi_0(a|x) = 
    \begin{cases} 
        1 & \text{if } a = a^* \\ 
        0 & \text{if } a \neq a^* \end{cases}
\end{equation}

Now, consider an evaluation policy $\pi$ which is also deterministic, based on a new scoring model $\text{score}_{\text{eval}}$:
\begin{equation}
    \pi(x) = a' = \underset{a \in A(x)}{\text{argmax}} (\text{score}_{\text{eval}}(a, x))
\end{equation}

When we apply the IPS estimator, we only consider the data points where an ad was shown ($a_i = a^*$). For these logged events, the importance weight $w_i$ can only take two values:
\begin{equation}
    w_i = \frac{\pi(a^*|x_i)}{\pi_0(a^*|x_i)} = \frac{\pi(a^*|x_i)}{1} = 
    \begin{cases} 
        1 & \text{if } \pi \text{ also chooses } a^* \\ 
        0 & \text{if } \pi \text{ chooses some } a' \neq a^* \end{cases}
\end{equation}

This leads to a critical limitation, as pointed out by previous works on off-policy evaluation in settings with limited support (e.g., \cite{bottou2013counterfactual, swaminathan2015self}). The IPS estimator simplifies to:
\begin{equation}
\hat V_{\mathrm{IPS}}(\pi)
= \frac{1}{n}\sum_{i=1}^n \mathbf{1}\{\pi(x_i)=\pi_0(x_i)\}\, r_i
\;\le\; \frac{1}{n}\sum_{i=1}^n r_i
= \hat V(\pi_0),
\end{equation}

\noindent\textbf{Under deterministic logging, the estimated value of the evaluation policy cannot exceed that of the logging policy}; equality holds only when $\pi$ agrees with $\pi_0$ on all impressions. Any disagreement is counted as a zero contribution, so potential improvements from the new policy are never credited, and $\hat V_{\mathrm{IPS}}(\pi)$ systematically underestimates the true value $V(\pi)$. This severe limitation necessitates a new approach that can look beyond the logged deterministic outcome and assign a meaningful, non-zero probability to an ad winning the auction.

\section{Proposed Method: DPM-OPE}\label{sec:proposed}
\begin{figure*}[h]
  \centering
  \includegraphics[trim=0cm 1.2cm 0cm 0cm,clip,width=0.75\textwidth]{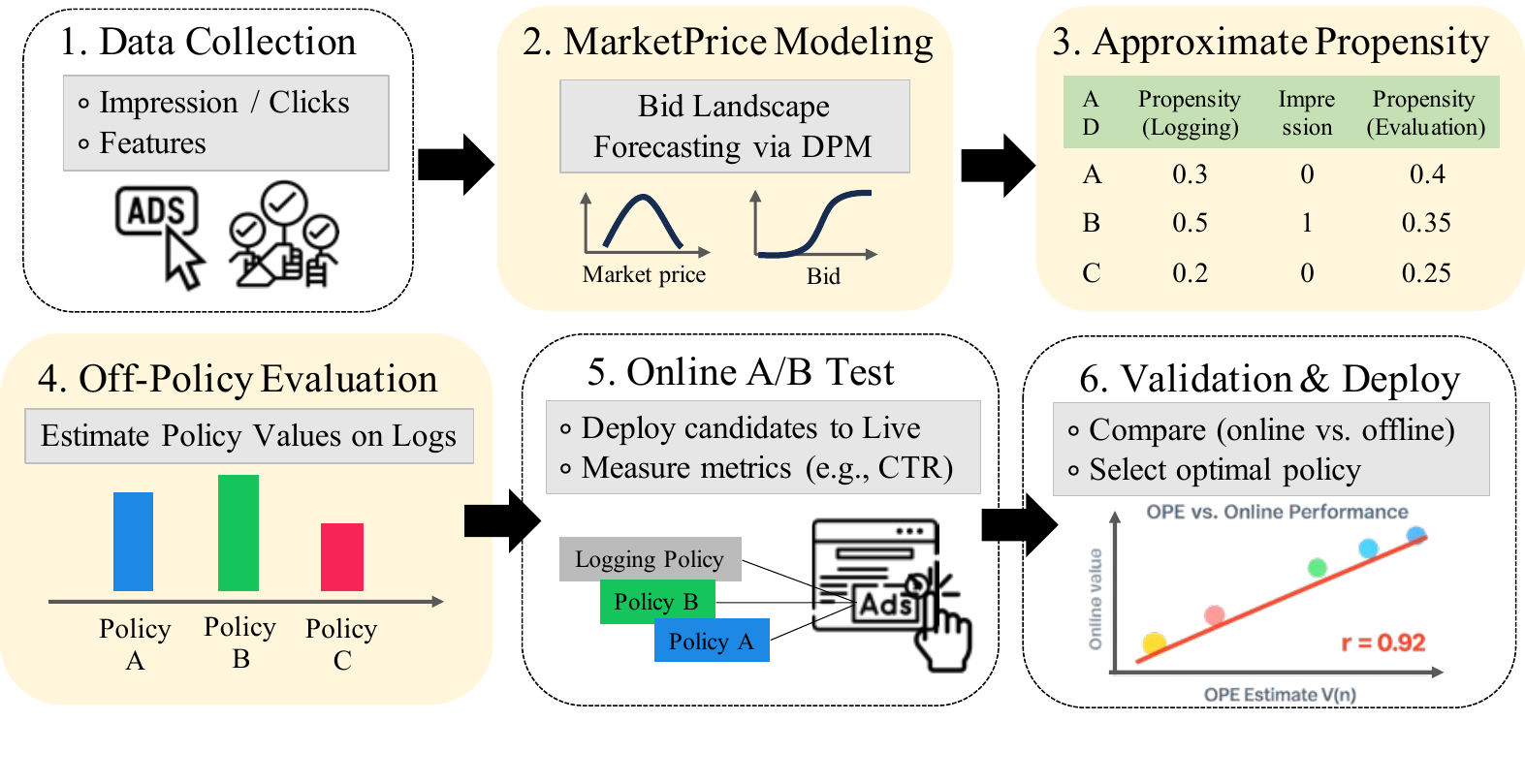}
  \caption{An overview of the end-to-end DPM-OPE framework: the technical steps 2, 3, and 4 (yellow-shaded) are described in Section \ref{sec:proposed}, and the remaining steps are described in Section \ref{sec:setup}.}
  \label{fig:figure_2}
  \Description{(1) We start with historical log data containing impressions, clicks, and features. (2) A Discrete Price Model (DPM) is trained on this data to model the market price landscape. (3) This model is used to derive an Approximate Propensity Score (APS) for each action. (4) The OPE estimators, now equipped with the APS, evaluate a set of candidate policies offline. (5) Top-performing candidates are then deployed in a live online A/B test. (6) The results from the online test are used to validate the accuracy of the offline evaluation, closing the loop.}
\end{figure*}

In the previous section, we established that the deterministic, winner-takes-all nature of online ad auctions renders standard OPE methods systematically biased.

To overcome this, we introduce \textbf{DPM-OPE}, a novel framework for Off-Policy Evaluation. Instead of attempting to stochastically approximate the deterministic policy directly, our approach fundamentally shifts the perspective to the inherent stochasticity of the \textit{auction environment} itself. The central concept is to model the competitive landscape that an advertisement encounters. By estimating the probability of an ad winning the auction given its score, we can derive a robust non-zero \emph{Approximate Propensity Score (APS)}. This APS serves as a valid substitute for the propensity score, enabling the use of powerful OPE estimators for reliable offline evaluation. Our framework is illustrated in Figure \ref{fig:figure_2}.

\subsection{A Shift in Perspective: Modeling the Auction Environment}

\begin{figure}[h!]
  \centering
  \includegraphics[trim=0cm 0.7cm 0.1cm 0cm,clip,width=0.4\textwidth]{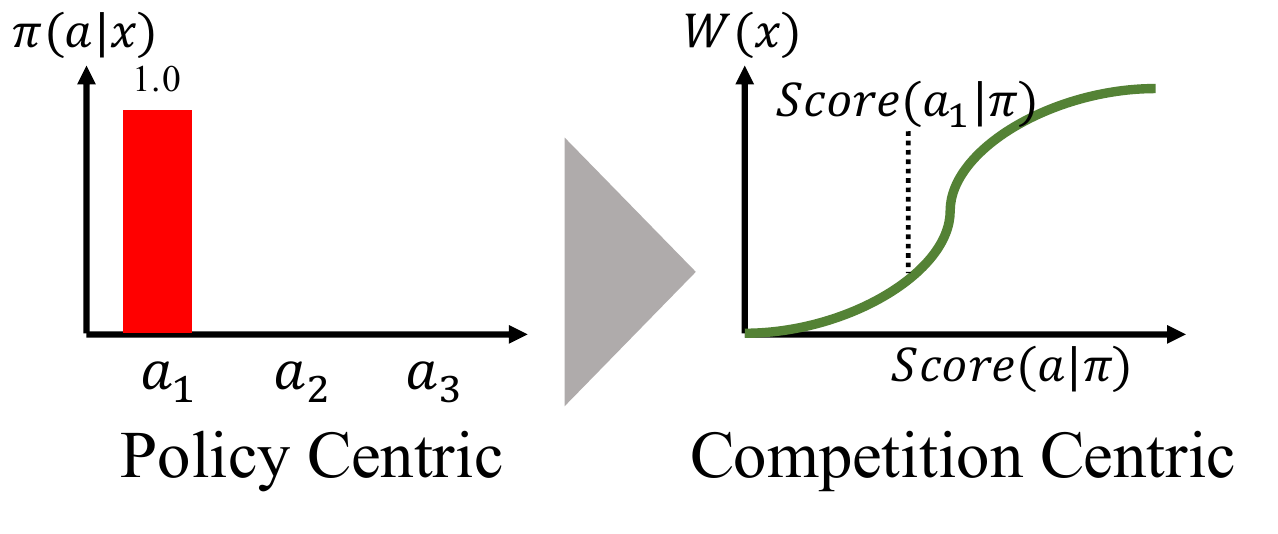}
  \caption{Reframing the Question: From Policy Choice to Winning Probability}
  \Description{Reframing the question: rather than modeling the policy’s choice $pi(a\mid x)$, we model the environment’s response—i.e., the probability that ad $a$ wins the auction. This shift captures market uncertainty and connects exposure with winning probability.}
  \label{fig:perspective_shift}
\end{figure}

Our perspective shifts from modeling the policy’s choice, $\pi(a|x)$, to modeling the competitive environment where that choice occurs. In an ad auction, an ad’s exposure is not a direct choice but the outcome of a competition. Therefore, the probability of an ad being shown equals its probability of winning that competition. This reframes the problem: instead of asking \emph{“What is the probability of the policy choosing ad $a$?”}, we ask, \textbf{\emph{“Given the competitive environment, what is the probability of ad $a$ winning?”}} (Figure~\ref{fig:perspective_shift}). This environment-centric view allows us to capture the market’s inherent uncertainty.

\subsection{Estimating the Market Price Distribution}

To estimate the winning probability, we first represent the entire set of competitors in an auction by a single random variable: the market price, defined as the second-highest score among all other ads \cite{wu2015predicting,zhang2016bid,wu2018deep,ren2019deep}.

\subsubsection{The Market Price and Winning Condition}
An ad $a_i$ with a score $s_i = \text{score}(a_i, x)$ wins an auction if its score is higher than that of all other competing ads. It is important to note that the formulation of $\text{score}(\cdot)$ varies depending on the auction mechanism. While it may correspond directly to a raw bid price in classic auctions, modern ad systems typically utilize a composite value, such as $\text{bid} \times \text{pCTR}$, to rank candidates.

In such environments, the determining factor for winning is this composite score. Consequently, the ``effective price'' one must beat to win the auction is the highest composite score among competitors. We therefore generalize the definition of market price $z$ as the highest score among all other competitors for a given impression:
\begin{equation}
    z = \max_{j \neq i} \{ \text{score}(a_j, x) \}
\end{equation}
The winning condition for ad $a_i$ is simply $s_i > z$. Since the set of competitors and their scores vary for each auction, the market price $z$ is a random variable. The probability of an ad with score $s$ winning is therefore $P(s > z)$, which is the cumulative distribution function (CDF) of the market price, often denoted as the winning function $W(s) = \int_0^s p(z) dz$. Our primary task is to model the probability distribution of the market price, $p(z)$.

\subsubsection{The Discrete Price Model (DPM)}
We employ a Discrete Price Model (DPM), a technique well-suited for this task and widely used in bid landscape forecasting \cite{ren2019deep,li2022arbitrary}. Since prices and scores in real-time bidding are inherently discrete due to finite precision, modeling them in a discrete space is a natural and effective choice \cite{liu2022real}.

We discretize the continuous score space into $L$ disjoint intervals, or bins, $V_l = (b_{l-1}, b_l]$ for $l=1, \dots, L$. This allows us to define the winning function $W(b_l)$ (the probability of winning with a score up to $b_l$) and the survival function $S(b_l)$ (the probability of losing with a score of $b_l$) over this discrete space:
\begin{equation}\label{eq:win-lose-def}
\begin{aligned}
W(b_l) & \doteq \text{Pr}(z < b_l) = \sum_{j < l} \text{Pr}(z \in V_j) \\
S(b_l) & \doteq \text{Pr}(z \geq b_l) = \sum_{j \geq l} \text{Pr}(z \in V_j)
\end{aligned}
\end{equation}
The probability of the market price $z$ falling into a specific interval $V_l$ is then given by:
\begin{equation}\label{eq:discrete-pdf-def}
p_l = \text{Pr}(z \in V_l) = S(b_{l-1}) - S(b_l)
\end{equation}


\subsubsection{Adaptive Binning}\label{sec:adaptive-binning}

Choosing the number of bins $L$ is crucial for balancing bias and variance in DPM. To ensure robust estimation, we determine the number of bins, $L$, \emph{adaptively from the data}. This process guarantees that the winning-rate estimate for each bin satisfies a predefined confidence interval width.

\paragraph{Setup.}
Let $n$ be the number of data points. We construct \emph{quantile bins} over the score $s=\mathrm{score}(a,x)$ so that each bin has approximately $n/L$ samples. Under a normal approximation to the binomial proportion $p$ in a bin, the two-sided $100\times(1-\alpha)\%$ confidence interval has half-width bounded by
\begin{equation}\label{eq:adaptive-accuracy}
z_{\alpha/2}\,\sqrt{\frac{p(1-p)}{\,n/L\,}} \;\le\; \varepsilon,
\end{equation}
where $z_{\alpha/2}$ is the standard normal quantile and $\varepsilon>0$ is the target accuracy.

\paragraph{Rule for $L$.}
We adopt the worst-case variance $p(1-p)\le \tfrac{1}{4}$ and tie the accuracy to the discretization granularity by $\varepsilon=\tfrac{1}{2L}$. Substituting into \eqref{eq:adaptive-accuracy} yields
\[
z_{\alpha/2}\,\sqrt{\frac{0.25\,L}{n}} \;\le\; \frac{1}{2L}
\quad\Longrightarrow\quad
L^{3} \;\le\; \frac{n}{\,z_{\alpha/2}^{2}\,}.
\]
For $\alpha=0.05$, $z_{\alpha/2}=1.96$ and $z_{\alpha/2}^{2}=3.8416$, which gives the closed-form sizing rule
\begin{equation}\label{eq:adaptive-L}
L^\star \;=\; \left\lfloor \left(\frac{n}{3.8416}\right)^{1/3} \right\rfloor.
\end{equation}

\paragraph{L as the maximum.}
The stochastic approximation introduces \emph{quantization bias}: using fewer bins mixes heterogeneous scores within a bin and flattens the winning function $W(s)$, thereby biasing the approximate propensities. The rule in \eqref{eq:adaptive-L} identifies the \emph{largest} $L$ that still satisfies the per-bin accuracy constraint; selecting $L=L^\star$ (subject to latency/memory caps) therefore minimizes discretization bias \emph{under the variance bound} implied by \eqref{eq:adaptive-accuracy}.

\paragraph{Segmentation rationale.}
We apply adaptive binning \emph{per segment} $g$ (e.g., per period in AuctionNet) because the score distribution can vary substantially across segments. To select a meaningful segmentation feature, we evaluate ANOVA explained variance on scores:
\[
\mathrm{SST} \;=\; \sum_{g}\sum_{i\in g} (s_i - \bar{s})^{2}, 
\qquad
\mathrm{SSB} \;=\; \sum_{g} n_g\,(\bar{s}_g - \bar{s})^{2},
\qquad
R^{2} \;=\; \frac{\mathrm{SSB}}{\mathrm{SST}},
\]
where $\bar{s}$ is the global mean, $\bar{s}_g$ is the group mean, and $n_g$ is the group size. We choose the segmentation that maximizes $R^{2}$, indicating that between-group differences explain a large portion of the total variance. After fixing the segmentation, we fit DPM and compute $L_g^\star$ per segment using \eqref{eq:adaptive-L}, then form quantile bins $V_\ell=(b_{\ell-1},b_\ell]$ within each segment.

\subsection{Approximate Propensity Score (APS)}

With the DPM framework established, we can derive a meaningful propensity score. While the cumulative winning probability $W(b_l)$ tells us the chance of beating the market, it does not represent the probability of a specific score being the winning one. For an importance weight, we need a value analogous to a probability mass function (PMF), not a cumulative distribution function (CDF).


To this end, we use the conditional winning probability, $h_l$, as our APS. This is defined as the probability that the market price falls within a specific interval $V_l$, given that it was not in any of the lower intervals:
\begin{equation}\label{eq:discrete-instant-win-func}
h_l = \text{Pr}(z \in V_l | z \geq b_{l-1}) = \frac{\text{Pr}(z \in V_l)}{\text{Pr}(z \geq b_{l-1})} = \frac{p_l}{S(b_{l-1})}
\end{equation}
The term $h_l$ represents the probability of just winning the auction with a score in the interval $V_l$. Therefore, we define the APS for an action $a$ with score $s \in V_l$ as:
\begin{equation}
    \pi_{\text{APS}}(a|x) \doteq h_l \quad \text{where } \text{score}(a,x) \in V_l
\end{equation}
This APS is a non-zero, granular probability that reflects the ad's likelihood of being the marginal winner, effectively creating the stochastic approximation we need.

We acknowledge that the accuracy of our framework is contingent on the quality of the estimated market price distribution, $P(z)$. Any misestimation in this step could potentially introduce bias into the final OPE result.

To mitigate this, we deliberately chose the DPM for its non-parametric flexibility, which allows it to capture complex, real-world distributions more faithfully than simpler parametric models. While our empirical results in Section 6 demonstrate that this approach is practically effective, we believe that developing more advanced distribution modeling techniques is a valuable direction for future research. Exploring methods that can even more precisely model the nuances of the market price would be a promising step toward further enhancing the accuracy of OPE in deterministic environments.

\subsection{DPM-OPE Estimator}
Having derived a valid, non-zero APS, we can now substitute it into a standard OPE estimator in place of $\pi_0(a|x)$ and $\pi(a|x)$. To mitigate the high variance often associated with the basic IPS estimator, we use the Self-Normalized Inverse Propensity Scoring (SNIPS) estimator \cite{swaminathan2015self}, which is known for its stability.

By replacing the problematic $\pi(a_i|x_i)$ with our derived $\text{APS}(a_i|x_i)$, we arrive at our final DPM-SNIPS estimator:
\begin{equation}
    \hat{V}_{DPM-SNIPS}(\pi) = \frac{\sum_{i=1}^n r_i \frac{\pi_{\text{APS}}(a_i|x_i)}{\pi_{\text{0 APS}}(a_i|x_i)}}{\sum_{i=1}^n \frac{\pi_{\text{APS}}(a_i|x_i)}{\pi_{\text{0 APS}}(a_i|x_i)}}
\end{equation}
This estimator allows us to reliably evaluate new policies $\pi$ using historical data from a deterministic logging environment, bridging the gap that previously made such offline evaluation impractical.

\section{Experiments}\label{sec:setup}

In this section, we present a series of experiments designed to validate the effectiveness and robustness of our proposed DPM-OPE framework. We aim to answer the following key questions:
\begin{enumerate}
    \item How accurately does DPM-OPE estimate the performance of new policies compared to the baseline in a controlled simulation environment?
    \item Does the effectiveness of DPM-OPE hold in a large-scale, real-world industrial setting?
    \item How sensitive is our method to its core hyperparameters?
\end{enumerate}

\subsection{Datasets \& Setup}

To thoroughly evaluate our framework, we utilized two distinct datasets: a public benchmark for controlled simulation and a large-scale industrial dataset for real-world validation.

\subsubsection{AuctionNet Benchmark}
Since there is no public benchmark specifically designed for OPE in ad auctions, we adapted \textbf{AuctionNet}\cite{su2024auctionnet}, a well-established benchmark for bidding decisions derived from a real-world advertising platform. To create an OPE testbed, we independently ran simulations for seven different bidding strategies, generating separate datasets for each.

We designated the ``PID'' bidding strategy as our logging policy ($\pi_0$) and the remaining six strategies (``BC'', ``BCQ'', ``CQL'', ``IQL'', ``OnlineLP'', ``TD3-BC'') as the evaluation policies ($\pi$) to be assessed via OPE. The ground truth performance for each evaluation policy was obtained from its corresponding simulation run. Key statistics for each simulation are presented in Table \ref{tab:auctionnet_stats}.

\begin{table}[h!]
\centering
\caption{Ground truth statistics for each policy simulation on the AuctionNet benchmark.}
\label{tab:auctionnet_stats}
\begin{tabular}{@{}lccc@{}}
\toprule
\textbf{Policy} & \textbf{Impressions} & \textbf{Conversions} & \textbf{CTR} \\ \midrule
pid ($\pi_0$)   & 9,679,017           & 10,317               & 0.001065      \\
bc              & 9,999,526           & 14,348               & 0.001434      \\
bcq             & 9,999,526           & 15,107               & 0.001510      \\
cql             & 9,999,526           & 14,730               & 0.001473      \\
iql             & 9,999,526           & 14,406               & 0.001440      \\
onlinelp        & 9,999,526           & 14,718               & 0.001471      \\
td3\_bc         & 9,999,526           & 14,639               & 0.001463      \\ \bottomrule
\end{tabular}
\end{table}

When applying our \emph{adaptive binning} strategy to the AuctionNet benchmark, the resulting number of bins $L$ ranged from approximately 48 to 50 across different segments (period), balancing estimation resolution with statistical reliability.

\subsubsection{Real-World Data}
To demonstrate the practical applicability of our framework, we collected a 2-week log from a large-scale online advertising platform at NAVER. This dataset contains the results of an online A/B test where two different CTR prediction models were deployed. The environment involves selecting one ad from a multitude of candidates for each ad slot. This extensive dataset provides a challenging and realistic testbed for evaluating our method's performance in an industrial setting.

\subsection{Evaluation Metrics}
\label{sec:evaluation_metrics}
A key goal in business contexts is not just to measure the absolute performance of a new model, but to accurately assess its \textit{relative improvement} over the current production model. Therefore, all our metrics are based on the CTR lift (or difference) of the evaluation policy relative to the logging policy.

\begin{table*}[h!]
\centering
\caption{Performance comparison on the AuctionNet benchmark and the 2-week real-world dataset. (↓ Lower is better, ↑ Higher is better)}
\label{tab:combined_results}
\begin{tabular}{@{}lcccc@{}}
\toprule
\multirow{2}{*}{\textbf{Metric}} & \multicolumn{2}{c}{\textbf{AuctionNet}} & \multicolumn{2}{c}{\textbf{Real-World (2-week)}} \\ \cmidrule(l){2-5}
 & \textbf{Parametric} & \textbf{DPM-OPE} & \textbf{Parametric} & \textbf{DPM-OPE} \\ \midrule
MDA (↑) & 100.0\% & 100.0\% & 78.6\% & \textbf{92.9\%} \\
RMSE (↓) & 25.585 & \textbf{4.870} & \textbf{1.187} & 1.197 \\
Pearson Correlation (↑) & \textbf{0.520} & -0.112 & 0.575 & \textbf{0.653} \\ \bottomrule
\end{tabular}
\end{table*}

\begin{itemize}
    \item \textbf{MDA (Mean Directional Accuracy)}: Measures the percentage of times our offline estimate correctly predicts the direction (improvement or degradation) of the CTR change. High MDA ensures stable day-to-day decision-making and \emph{minimizes the risk of deploying underperforming models}.
    \item \textbf{RMSE (Root Mean Square Error)}: Measures the average of the squares of the error between the ground truth CTR lift and the lift estimated by OPE. We report these in percentage points (\%-points) for intuitive comparison.
    \item \textbf{Pearson Correlation}: Measures the linear relationship between the daily performance trends of the ground truth and the OPE estimates, indicating how well our offline metric tracks online performance fluctuations.
\end{itemize}

\textbf{Predicting the correct direction of performance change} (i.e., whether CTR will improve or degrade) is more important than minimizing numerical error or maximizing correlation. Accordingly, we prioritize our metrics in the following order: \textbf{MDA} $>$ \textbf{RMSE} $>$ \textbf{Pearson Correlation}.

\subsection{Baseline}
We compare our DPM-OPE framework against a strong, domain-aware baseline. Before introducing it, we first justify the exclusion of a more generic class of methods for deterministic policies: kernel-based smoothing~\cite{kallus2020doubly, lee2022local, lee2024kernel}. This common technique "relaxes" a deterministic choice by placing a continuous probability kernel (e.g., Gaussian) around the chosen action, thereby creating a synthetic stochastic policy to enable OPE.

However, we argue that this generic approach is fundamentally misaligned with the mechanics of our ad auction problem for two primary reasons. First, the action space consists of a discrete set of competing ad candidates, which lacks the natural distance metric that continuous kernels rely on. Second, and more critically, such methods are agnostic to the underlying auction dynamics. They ignore the complex, often multimodal market price distribution that truly governs the winning probability.

Instead of artificially smoothing the action space, our work posits that a more principled approach is to directly model the source of the environmental stochasticity, the market price. Therefore, we select a baseline that adheres to this principle:
\begin{itemize}
    \item \textbf{Parametric-OPE}: This baseline assumes that the market price distribution follows a known parametric form. For each auction, we fit the observed bid prices of competing ads to several distributions (Normal, Log-Normal, Beta, Gamma, Exponential) and select the one with the best fit. The APS is then calculated from the Cumulative Distribution Function (CDF) of the chosen parametric distribution.
\end{itemize}


\subsection{Analysis of Framework Components}

To validate the specific design choices within our DPM-OPE framework, we designed a series of analytical experiments. This section outlines the setup for these studies, which investigate the impact of two key components: (1) the DPM's binning strategy and (2) the final OPE estimator. The results of these comparisons will be presented in detail in Section 6.

\subsubsection{Impact of Binning Strategy}
The discretization of the score space into bins is a core component of the DPM. To understand its impact, we will compare two distinct quantile-based strategies:
\begin{itemize}
    \item \textbf{Static Binning}: This approach uses quantile binning to ensure each bin contains an equal number of samples, but the total number of bins ($L$) is a fixed hyperparameter (e.g., 100, 1000). While effective, the choice of $L$ requires manual tuning and may not be optimal across different datasets.
    \item \textbf{Adaptive Binning (Our Method)}: This strategy determines the number of bins $L$ adaptively from the data. The goal is to set $L$ such that the winning-rate estimate within each bin achieves a target statistical precision at a prescribed confidence level. This removes the need to manually specify the number of bins.
\end{itemize}

\subsubsection{Choice of OPE Estimator}
While our primary contribution is the derivation of the Approximate Propensity Score (APS), the choice of the final estimator that consumes this score is crucial for stability. We will compare three variants based on the Inverse Propensity Scoring (IPS) family:
\begin{itemize}
    \item \textbf{IPS}: The standard, unbiased estimator. However, it is known to suffer from high variance, especially if some importance weights are large.
    \item \textbf{SNIPS (Self-Normalized IPS)}: This estimator normalizes the weights, which is known to reduce variance and improve stability compared to the standard IPS.
    \item \textbf{Capped SNIPS (Our Final Choice)}: Even with SNIPS, a few outlier events with extremely small APS values can lead to excessively large importance weights, destabilizing the final estimate. To address this, we employ a common technique of capping the importance weights at the 99th percentile of their distribution. This trades a small amount of potential bias for a significant reduction in variance.
\end{itemize}

\section{Results}

\subsection{Simulation Results: AuctionNet}\label{sec:simulation_results}
\begin{figure}[h!]
    \centering
    \includegraphics[width=\linewidth]{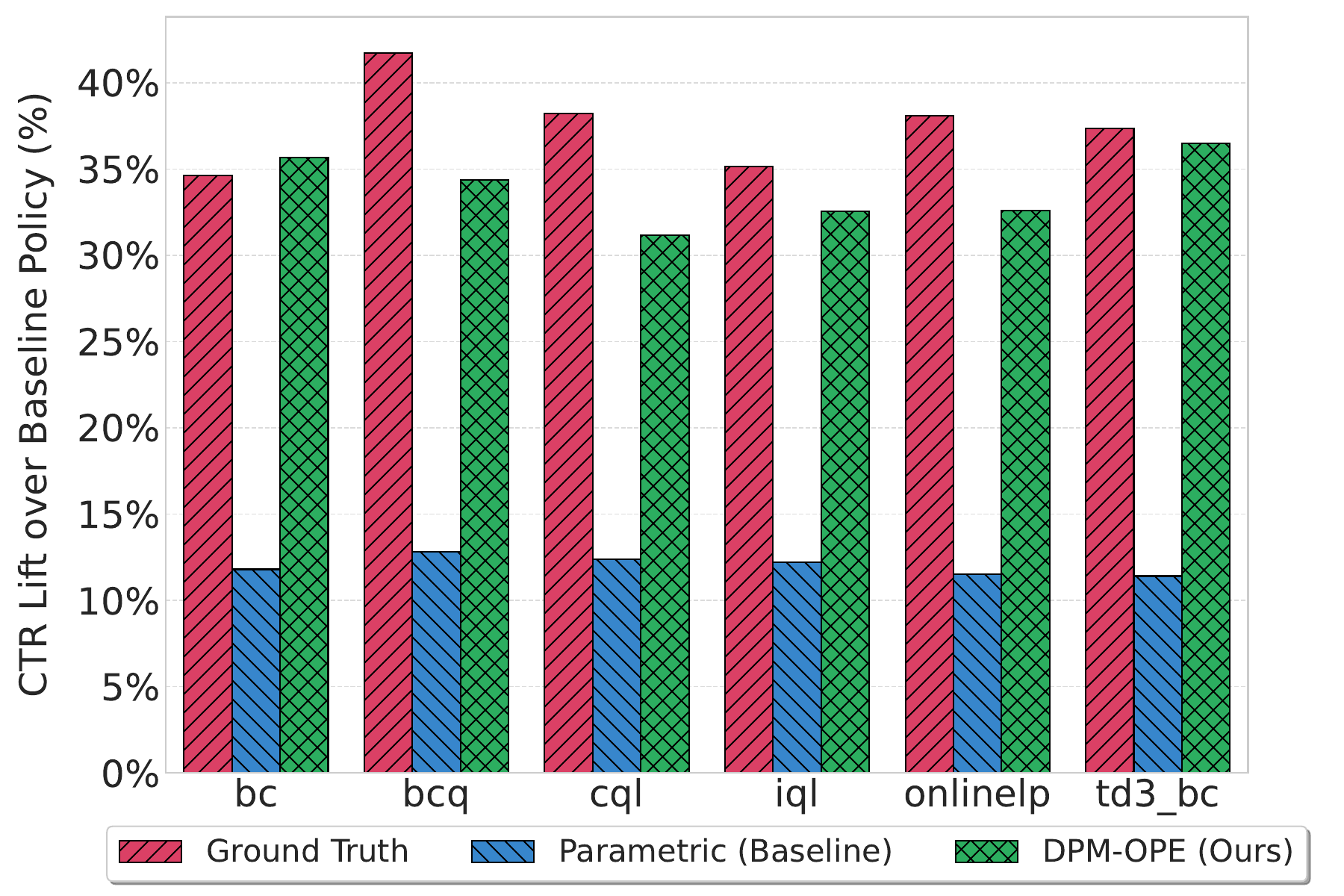}
    \caption{Comparison of estimated CTR lift vs. ground truth on AuctionNet. DPM-OPE's estimates (green bars) are consistently closer to the ground truth (red bars) than the parametric baseline's estimates (blue bars).
    }
    \label{fig:auctionnet_plot}
    \Description{This bar chart displays the CTR lift of six different evaluation policies relative to the PID logging policy. It compares the ground truth lift obtained from simulation A/B tests against the estimates from our DPM-OPE framework and the parametric baseline.}
\end{figure}

Table \ref{tab:combined_results} and Figure \ref{fig:auctionnet_plot} summarize the performance on the AuctionNet benchmark. Our DPM-OPE method demonstrates a clear advantage in estimation accuracy, achieving a substantially lower RMSE than the parametric baseline. This indicates that DPM-OPE predicts the true CTR lift with significantly higher precision, which is the primary goal of offline evaluation.

While the parametric baseline shows a higher Pearson Correlation, this metric should be interpreted with caution in this specific simulation. The ground truth CTR differences between several evaluation policies are not statistically significant (e.g. \textit{BC vs IQL or BC vs TD3-BC}). Consequently, a high correlation derived from the ranking of these near identical outcomes is less indicative of true performance than the absolute error metrics, where our method clearly excels.

\subsection{Real-world Online AB Testing: NAVER Platform}

\begin{figure}[h!]
    \centering
    \includegraphics[width=\linewidth]{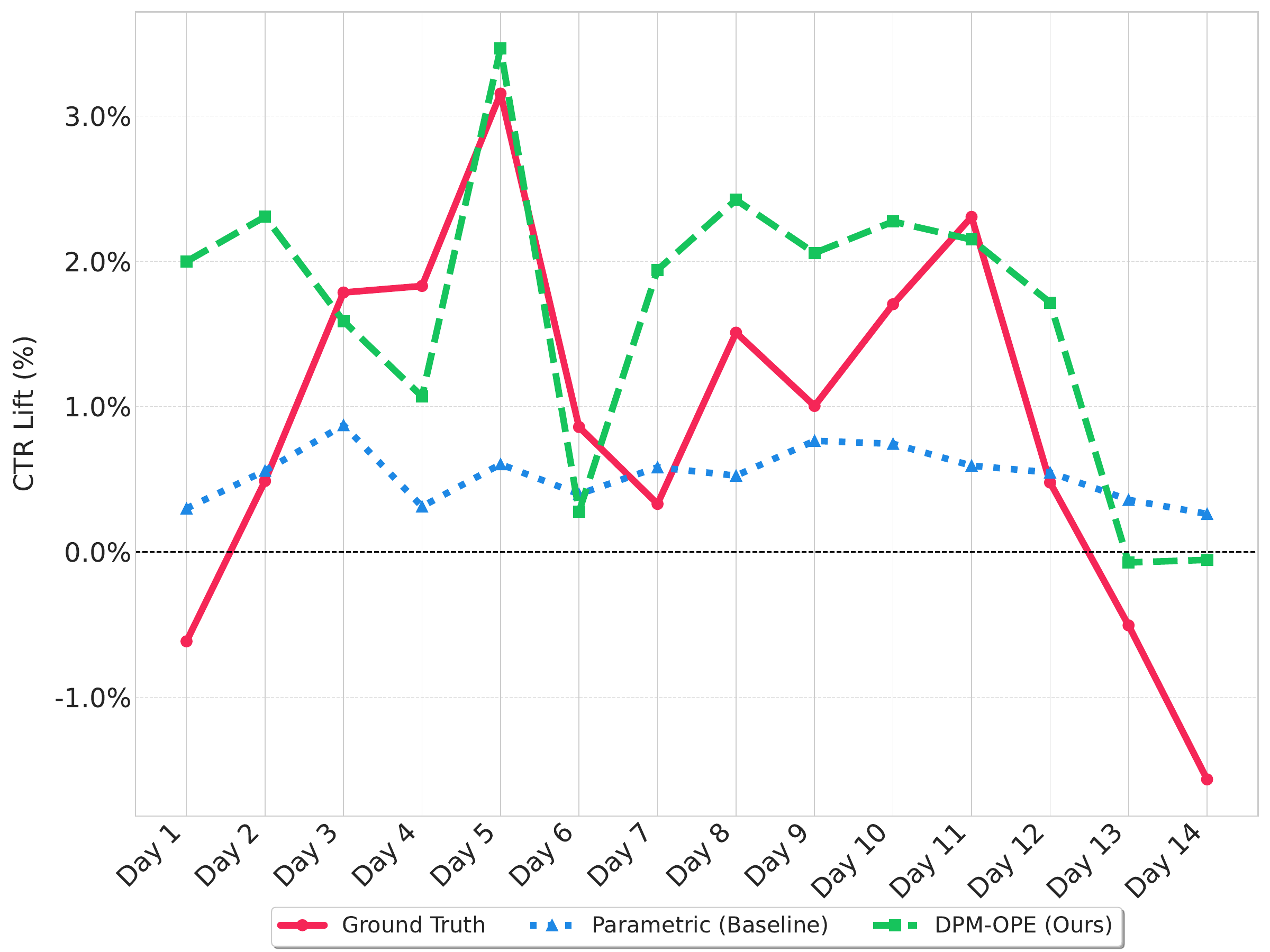}
    \caption{Daily trend comparison on real-world data. The DPM-OPE estimates (green squares) closely follow the ground truth online A/B test results (red circles), while the baseline (blue triangles) fails to capture the trend.}
    \label{fig:realworld_plot}
    \Description{This line graph compares the daily CTR lift estimates from DPM-OPE and the parametric baseline against the ground truth from an online A/B test. From a business perspective, correctly identifying the direction of change (MDA) is critical for decision-making. The baseline method is unreliable as it incorrectly suggests a positive lift on most days, whereas our DPM-OPE method demonstrates high reliability with a much stronger MDA.}
    \vspace{-1em}
\end{figure}

The results on the 2-week real-world dataset, shown in Table \ref{tab:combined_results} and Figure \ref{fig:realworld_plot}, further highlight the practical advantages of our framework. While the parametric baseline achieved a marginally lower RMSE, the difference between the RMSE of the parametric baseline and that of our DPM-OPE method was found to be statistically insignificant (\textit{p=0.9751} via a paired t-test at a 95 \% confidence level). As mentioned in Section ~\ref{sec:evaluation_metrics}, metrics that directly inform business decisions are often more critical in real-world environments. Specifically, DPM-OPE achieved a much higher MDA (92.9\% vs. 78.6\%), a 14.3 percentage point improvement. This demonstrates a superior ability to make correct directional predictions about whether a new model will improve or harm performance, which is paramount for practical deployment.

Furthermore, the significantly higher Pearson Correlation (0.653 vs. 0.575) shows that our method more reliably tracks the day-to-day fluctuations of the online A/B test, as visualized in Figure \ref{fig:realworld_plot}. For industrial applications, this trend-following capability is often more valuable than marginal gains in average error, as it provides a more trustworthy signal for monitoring and analysis. These results strongly suggest that DPM-OPE offers a more robust and practical solution for real-world offline evaluation.

\begin{figure}[H]
  \centering
  \includegraphics[trim=0cm 0.1cm 0.1cm 0cm,clip,width=0.45\textwidth]{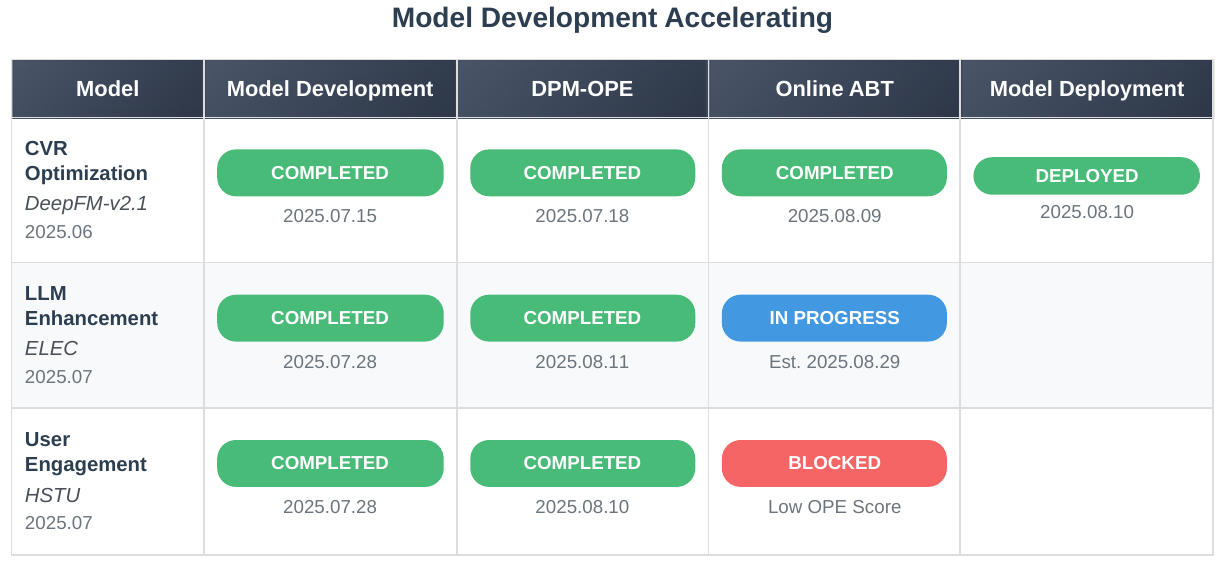}
  \caption{Industrial adoption of DPM-OPE showing its role in accelerating model development.}
  \Description{This figure illustrates a real-world industrial adoption case in which DPM-OPE is applied before online A/B testing. By accurately filtering out underperforming models offline, the process reduces the number of costly online experiments, shortens the development cycle, and enables quicker release of high-performing models in a production environment.}
  \label{fig:model_dev_accelerating}
\end{figure}

Beyond controlled experiments, DPM-OPE has seen industrial adoption in a large-scale advertising platform, where it is routinely used to pre-screen candidate models offline before any online A/B testing. 
In one case, a third candidate model with a low OPE score bypassed the ABT stage entirely, avoiding the cost of an unnecessary online experiment. 
This practice has measurably reduced the volume of costly online experiments and accelerated the overall model development cycle, further validating the framework's practical value in real-world settings (Figure~\ref{fig:model_dev_accelerating}).

\subsection{Ablation Studies}
To validate the specific design choices within our DPM-OPE framework, we conducted further analysis on the AuctionNet dataset. We investigated the impact of two key components: the DPM's binning strategy and the final OPE estimator. As mentioned in Section ~\ref{sec:simulation_results}, the Pearson correlation for AuctionNet was not statistically significant, so it was excluded from the ablation studies.

\subsubsection{Impact of Binning Strategy}

\begin{table}[h!]
\vspace{-1em}
\centering
\caption{Metrics per binning strategy on AuctionNet.}
\label{tab:auctionnet_bin_strategy}
\begin{tabular}{@{}lccc@{}}
\toprule
\textbf{Method} & \textbf{RMSE (↓)} & \textbf{MDA (↑)} \\ \midrule
Bin=100    & 5.509 & 100 \%  \\
Bin=1,000  & 6.461 & 100 \% \\
Bin=10,000 & 18.504 & 100 \% \\
\textbf{Bin=Adaptive}  & \textbf{4.870} & 100 \% \\ \bottomrule
\end{tabular}
\vspace{-0.5em}
\end{table}

\begin{figure}[h!]
    \centering
    \includegraphics[width=\linewidth]{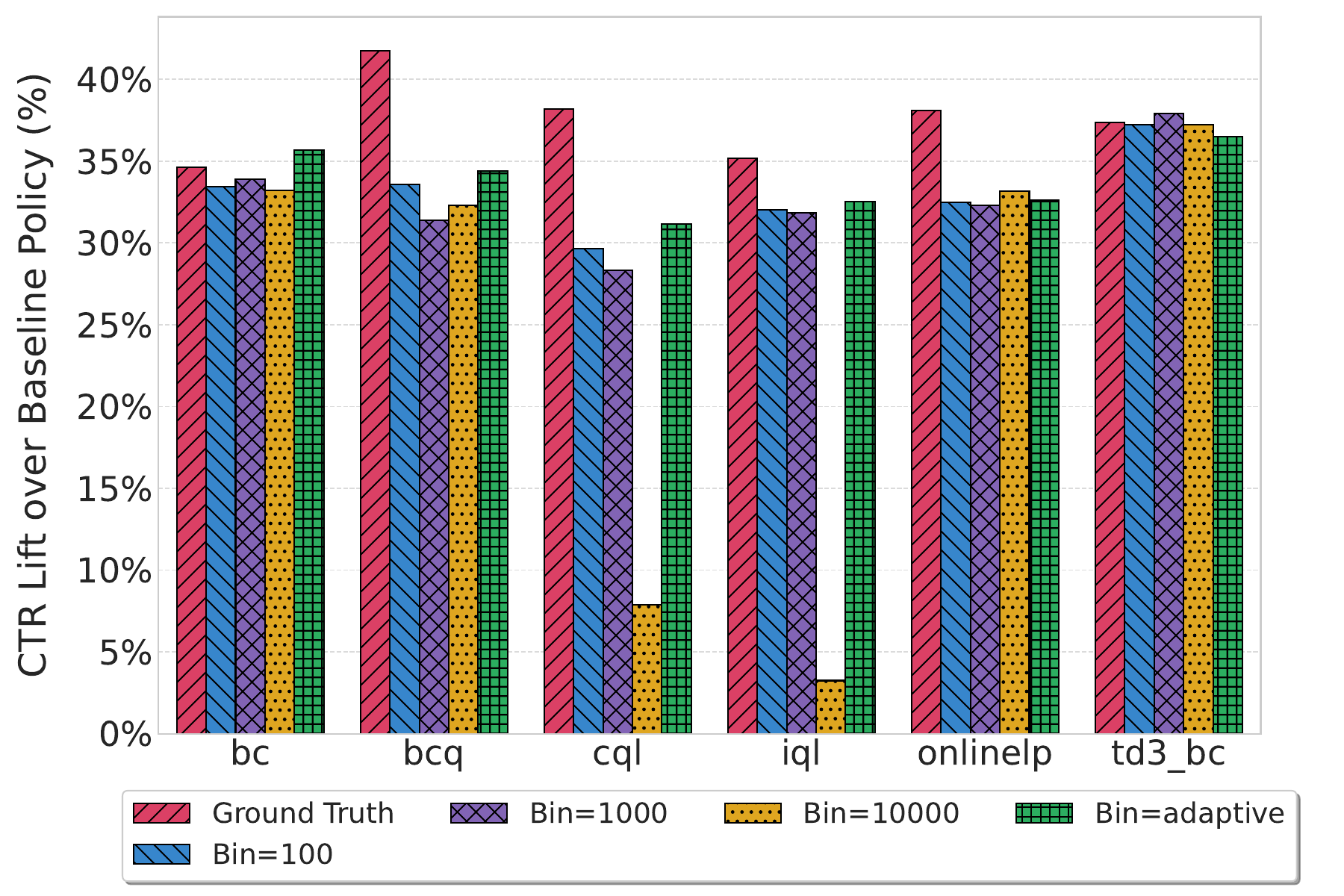}
    \caption{Performance comparison of different binning strategies on AuctionNet.}
    \label{fig:auctionnet_bin_strategy}
    \Description{This study compares our adaptive binning strategy against static quantile binning with a fixed number of bins (100, 1,000, and 10,000) to validate its effectiveness. Adaptive binning is highly efficient as it achieves robust performance without requiring manual tuning of the bin count for different datasets.}
\end{figure}

Table \ref{tab:auctionnet_bin_strategy} and Figure \ref{fig:auctionnet_bin_strategy} compare our proposed adaptive binning strategy against static quantile binning with a fixed number of bins (100, 1,000, and 10,000). Our adaptive binning approach achieves the lowest RMSE, providing the most stable and accurate estimates. This result validates our choice of this method for robust performance without manual hyperparameter tuning.

\subsubsection{Choice of OPE Estimator}

\begin{table}[h!]
\vspace{-0.5em}
\centering
\caption{Metrics per OPE estimator on AuctionNet.}
\label{tab:auctionnet_estimator}
\begin{tabular}{@{}lccc@{}}
\toprule
\textbf{Estimator} & \textbf{RMSE (↓)} & \textbf{MDA (↑)} \\ \midrule
IPS    & 355.532 & 100 \% \\
SNIPS  & 20.448 & 100 \%  \\
\textbf{Capped SNIPS}  & \textbf{4.870} & 100 \% \\ \bottomrule
\end{tabular}
\vspace{-0.5em}
\end{table}

\begin{figure}[h!]
    \centering
    \includegraphics[width=\linewidth]{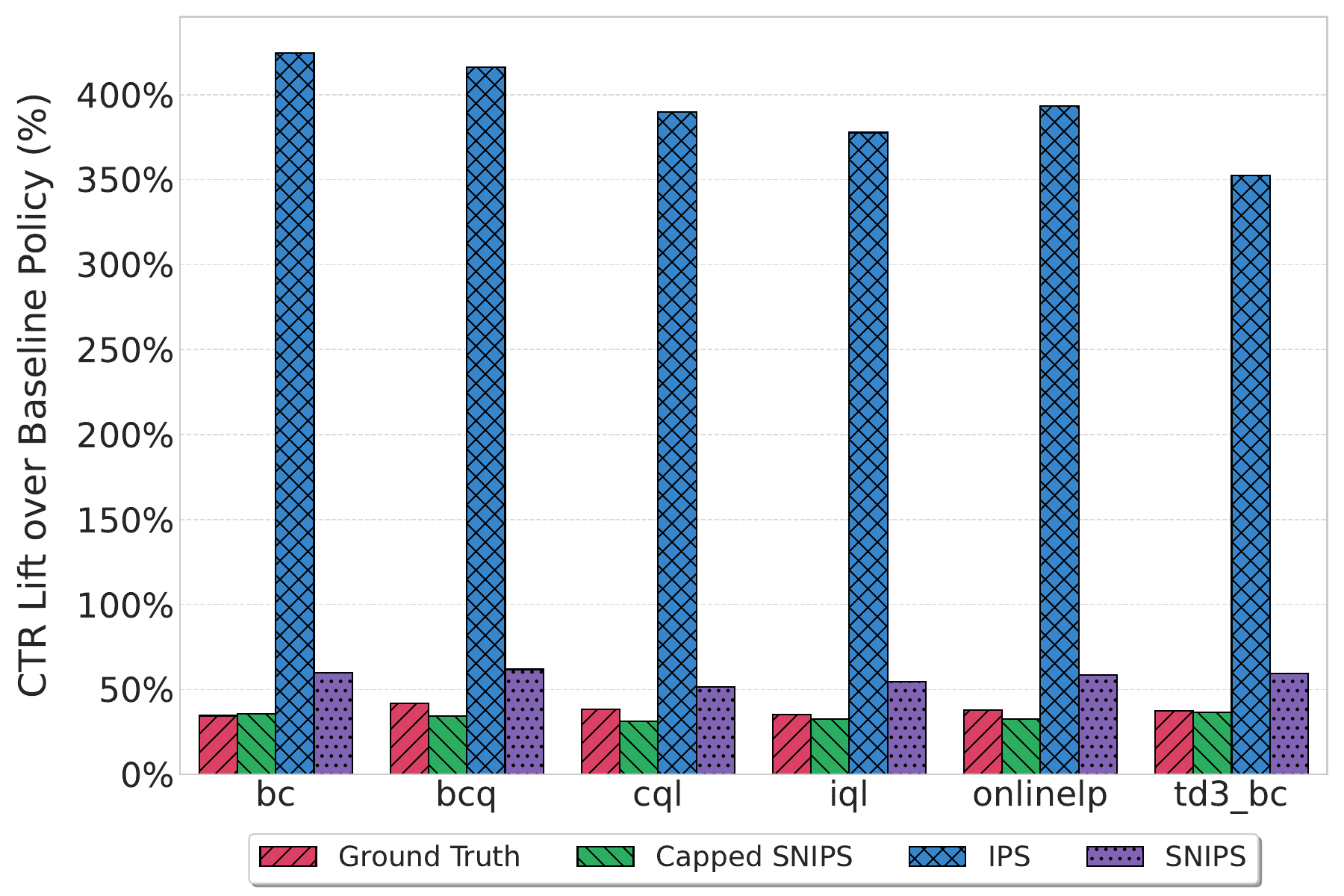}
    \caption{Performance comparison of different OPE estimators on AuctionNet.}
    \label{fig:auctionnet_estimator}
    \Description{This experiment compares the performance of IPS, SNIPS, and our final choice, Capped SNIPS. It highlights the critical importance of capping the importance weights to control their high variance, a common and significant challenge in both benchmark and real-world industrial data.}
\end{figure}

Table \ref{tab:auctionnet_estimator} and Figure \ref{fig:auctionnet_estimator} illustrate the importance of the estimator choice. The standard IPS estimator suffers from extremely high variance, making its estimates unusable in practice. Although SNIPS offers a significant improvement, its error rates remain high. Our final choice, Capped SNIPS, significantly reduces the variance by clipping outlier importance weights. This results in the lowest RMSE by a large margin, confirming that capping is an essential step to achieve a stable and reliable OPE framework.

\section{Discussion and Future Directions}
In this paper, we addressed the critical challenge of performing Off-Policy Evaluation in deterministic, winner-takes-all ad auction environments, a setting where standard estimators fail due to the zero propensity problem.
We introduced DPM-OPE, a framework that transforms the deterministic environment into a stochastic one by modeling the market price distribution, producing a robust Approximate Propensity Score (APS).

Our contributions are both methodological and practical:
\begin{itemize}
    \item \textbf{the first validated application of DPM to deterministic-auction OPE}, by repurposing the DPM from bid landscape forecasting for OPE.  Unlike its original use for future bid optimization, we employ DPM to estimate the probability of \textit{past} actions, enabling OPE where it was previously inapplicable. To our knowledge, this is the first comprehensive, validated solution to the zero propensity problem in online advertising logs.
    \item \textbf{strong empirical validation on both public benchmark and industrial AB test}, achieving higher directional accuracy and better trend tracking than the baseline. 
    \item This work is not only a \textbf{validated and deployable solution for the advertising industry}, but also a \textbf{truly practical contribution} that can be easily and flexibly adapted to other industries, operating under deterministic policies, such as logistic routing and deterministic ranking systems.
\end{itemize}

Despite its effectiveness, DPM-OPE has the following limitations and possible directions for future work:
\begin{itemize}
    \item The current DPM treats the market price distribution as dependent solely on aggregated scores, ignoring other contextual features (e.g., ad type, user attributes). More advanced, feature-aware models such as Deep Landscape Forecasting (DLF) or Arbitrary Distribution Modeling (ADM), could improve personalization and accuracy.
    \item The framework’s accuracy hinges on how well the market price distribution $P(z)$ is estimated. Misestimation can introduce bias into the final OPE results. While DPM mitigates this through non-parametric flexibility, further theoretical study is needed to quantify and control error propagation.
    \item Although adaptable to other deterministic-policy domains, applying DPM-OPE outside ad auctions may require domain-specific calibration, particularly for defining an equivalent to ``market price.'' Also, our approach focuses on propensity correction rather than reward modeling. Combining APS estimation with model-based or doubly robust estimators could potentially reduce variance further.
\end{itemize}



\bibliographystyle{ACM-Reference-Format}
\bibliography{references}

\end{document}